\documentclass{ecai} 

\usepackage{latexsym}
\usepackage{amssymb}
\usepackage{amsmath}
\usepackage{amsthm}
\usepackage{booktabs}
\usepackage{enumitem}
\usepackage{graphicx}
\usepackage{tabularx}
\usepackage{color}
\usepackage{multirow}
\usepackage{array}
\usepackage{ragged2e}
\usepackage{hyperref}
\usepackage{caption}
\usepackage{float}
\usepackage{placeins}
\usepackage{adjustbox}

\newcommand{\BibTeX}{B\kern-.05em{\sc i\kern-.025em b}\kern-.08em\TeX}

\begin{document}


\begin{frontmatter}


\title{GPT and Prejudice: A Sparse Approach to Understanding Learned Representations in Large Language Models}

\author[A]{\fnms{Mariam}~\snm{Mahran}\orcid{0009-0003-0568-0172}\thanks{Corresponding Author. Email: mariam.mahran@htw-berlin.de.}\footnote{Equal contribution.}}
\author[A]{\fnms{Katharina}~\snm{Simbeck}\orcid{0000-0001-6792-461X}\thanks{Corresponding Author. Email: katharina.simbeck@htw-berlin.de.}\footnotemark[1]}

\address[A]{HTW Berlin University of Applied Sciences, Treskowallee 8, 10318 Berlin, Germany}


\begin{abstract}
Large Language Models (LLMs) are trained on massive, unstructured corpora, making it unclear which social patterns and biases they absorb and later reproduce. Existing evaluations typically examine outputs or activations, but rarely connect them back to the pre-training data. We introduce a pipeline that couples LLMs with sparse autoencoders (SAEs) to trace how different themes are encoded during training. As a controlled case study, we trained a GPT-style model on 37 nineteenth-century novels by ten female authors, a corpus centered on themes such as gender, marriage, class, and morality. By applying SAEs across layers and probing with eleven social and moral categories, we mapped sparse features to human-interpretable concepts. The analysis revealed stable thematic backbones (most prominently around gender and kinship) and showed how associations expand and entangle with depth. More broadly, we argue that the LLM+SAEs pipeline offers a scalable framework for auditing how  cultural assumptions from the data are embedded in model representations.

\end{abstract}

\end{frontmatter}


\section{INTRODUCTION}

Modern Large Language Models (LLMs) are trained on massive, heterogeneous datasets combining web pages, books, code repositories, and user-generated content \cite{kim2025detectingtrainingdatalarge}. A well-known example is Common Crawl, which supplied much of the training data for GPT-3 \cite{brown2020languagemodelsfewshotlearners}. The sheer size and lack of structure of such corpora make them impossible to audit manually. As a result, the structures, biases, and themes embedded in the data remain opaque, despite their influence on model behavior \cite{10.1145/3442188.3445922}.

Research on bias in NLP and LLMs has largely targeted either model outputs or internal representations \cite{gallegos2024biasfairnesslargelanguage,li2024surveyfairnesslargelanguage}. Yet these approaches often fail to connect observed biases back to their origins in pretraining data \cite{thaler2024farbiasgo}. Some recent work has begun to explore this connection, but progress is limited due to inaccessibility of commercial training corpora \cite{köksal2023languageagnosticbiasdetectionlanguage,orgad-belinkov-2022-choose,thaler2024farbiasgo}. Without visibility into the data itself, it remains difficult to understand how models internalize societal narratives and assumptions. 

To bridge this gap, we propose coupling LLMs with sparse autoencoders (SAEs) as a pipeline to uncover conceptual structures in training data. SAEs have recently emerged within \textit{mechanistic interpretability} as a promising tool for exposing the factors that shape model behavior \cite{cunningham2023,hindupur2025projectingassumptionsdualitysparse,jing2025sparseautoencoderinterpretslinguistic,lou2025saevinterpretingmultimodalmodels, Simon2024.11.14.623630,paulo2024automaticallyinterpretingmillionsfeatures, simbeck2025saes}. By mapping dense activations into a sparse latent space, they help disentangle overlapping signals revealing how features are encoded. SAEs make it possible to recover high-level, interpretable features that reflect the social patterns, themes, and biases embedded in the underlying corpus.

As a case study, we trained a GPT-style model on a curated collection of novels by ten female authors from the late eighteenth and nineteenth centuries, known for their sustained engagement with themes of gender, marriage, class, and morality. By applying SAEs across all layers, we trace how these constructs appear in the model’s internal activations. While the model itself is modest in scale and not optimized for benchmarks, it offers a controlled setting to examine the link between data and representation.

Beyond this, our pipeline provides a practical method for mapping model states to human-interpretable categories. This is not a new technique, but it illustrates how SAEs can serve as a lens onto the data. Demonstrated on a compact, historically grounded corpus, the approach is readily adaptable to larger and noisier datasets, enabling the recovery of hidden structures, patterns, and biases that LLMs inherit from their training material.


\begin{figure*}[t]
    \centering
    \includegraphics[width=\textwidth]{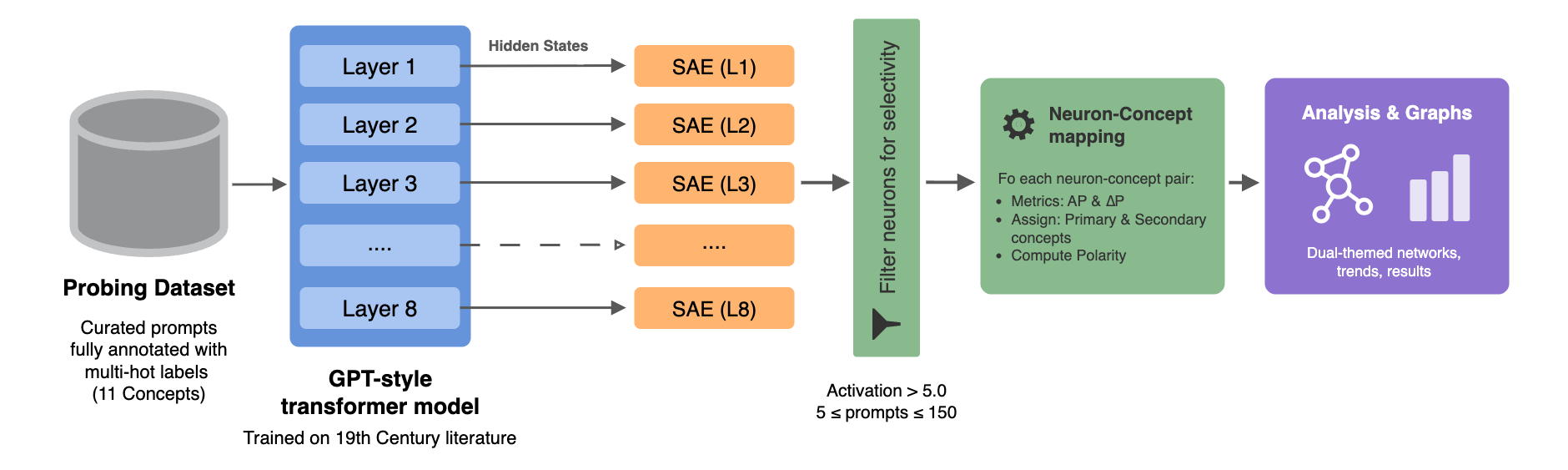}
    \vspace{1ex}
    \caption{Overview of the probing pipeline. Curated prompts annotated with 11 concepts are passed through a GPT-style model trained on nineteenth-century novels. Hidden states from each layer are encoded by sparse autoencoders (SAEs), then filtered for selectivity. For each retained neuron-concept pair, metrics (AP, $\Delta P$), primary/secondary assignments, and polarity are computed, forming the basis for our analyses.}
    \label{fig:pipeline_illust}
    \vspace{3ex}
\end{figure*}

\section{RELATED WORK}

\subsection{The Imprint of Pretraining Data}

As LLMs have become increasingly powerful and more fluent, there has been a renewed focus on understanding their internal mechanisms, particularly how the data they are trained on influences their behavior. Recent work suggests that LLMs may encode structured world knowledge, as certain features emerge consistently across different models \cite{chughtai2023toymodeluniversalityreverse,baek2024generalizationstarvationhintsuniversality}. However, this stands in contrast to the “stochastic parrot” view, which argues that models merely reproduce patterns in their corpora rather than truly “understanding” language \cite{10.1145/3442188.3445922}. This perspective reframes model behavior not as a window into intelligence, but as a mirror of the data these models consume. The latter argument is reinforced by recent findings that LLMs often succeed on surface-level tasks but fail at deeper reasoning, suggesting much of their apparent intelligence reduces to memorized correlations \cite{yu2025stochasticparrotllmsshoulder}. 

This influence of data extends to any learned bias. Research shows that LLMs replicate and even amplify stereotypes related to gender, race, religion, and social class \cite{gallegos2024biasfairnesslargelanguage,köksal2023languageagnosticbiasdetectionlanguage,orgad-belinkov-2022-choose, mahran2025multilingualpipeline, simbeck2025saes}. For instance, one study found consistent social bias in code generation outputs \cite{Ling_Rabbi_Wang_Yang_2025}. Another found that larger models amplify gender bias more strongly, suggesting scale exacerbates representational harm \cite{dong2024disclosuremitigationgenderbias}.

Nonetheless, these studies often treat models as black boxes, addressing symptoms rather than sources. Since LLMs are shaped by their training data, the data itself must be examined. Recent work on membership inference attacks (MIAs) show that individual training examples leave detectable traces \cite{kim2025detectingtrainingdatalarge}, providing direct evidence of the imprint of pretraining corpora. Similarly, analyses of large datasets reveal clear presence of occupational stereotypes, such as gendered job associations, that reappear (and sometimes intensify) in model outputs \cite{thaler2024farbiasgo}. Efforts like prompt engineering or instruction tuning offer only limited mitigation.


A larger issue is that most modern commercial LLMs rely on massive corpora scraped indiscriminately from the internet. These collections include unexpected sources such as patents, military websites, and even machine-generated content \cite{dodge2021documentinglargewebtextcorpora}. Their lack of structure and immense scale place them beyond the reach of manual auditing, leaving biases and hidden conceptual patterns undiscovered until after training. These concerns underline our motivation to develop tools that help recover and interpret the conceptual structures encoded by LLMs from their data.

\subsection{Sparse Autoencoders}

A key challenge with traditional interpretability approaches (e.g. attention weight visualization or probing individual neurons) is the problem of \textbf{superposition} where multiple overlapping features are compressed into the same neuron \cite{elhage2022toymodelssuperposition,bricken2023monosemanticity}. This makes it unclear whether activations reflect a relevant concept or interference from other signals.

Sparse autoencoders (SAEs) address superposition by projecting activations into a larger latent space while enforcing \textbf{sparsity}, so that only a few units fire for each input \cite{cunningham2023,bricken2023monosemanticity}. An SAE is a simple neural network typically constructed with two fully connected layers with a non-linear activation. SAEs are trained to reconstruct the original activations from a layer within the LLM. The encoder maps these activations into a larger latent space, while enforcing sparsity, and the decoder reassembles them. Sparsity is usually imposed through techniques like L1 regularization \cite{cunningham2023,LI2021107003}, top-k masking \cite{makhzani2014ksparseautoencoders}, or thresholding functions such as JumpReLU \cite{rajamanoharan2024jumpingaheadimprovingreconstruction}.


Recent studies confirm the utility of SAEs in the interpretability of LLMs. They were successfully used to identify linguistic structures aligned with syntax and semantics \cite{cunningham2023}, linguistic patterns, such as phonetic, syntactic, and pragmatic features distributed across layers \cite{jing2025sparseautoencoderinterpretslinguistic}, and even domain-specific signals such as protein functions in biological models \cite{Simon2024.11.14.623630}.

Building on these findings, we shift the focus from interpreting the model in isolation to using it as a lens onto the data it was trained on. By coupling SAEs with a language model, we can peek back into the corpus that shaped it.


\section{METHODOLOGY}
\label{sec:methodology}

To examine how thematic patterns are encoded in language models, we trained a GPT-style transformer on a curated corpus of nineteenth-century literature. Hidden states from its layers were analyzed with sparse autoencoders to probe the structure of learned representations. This section outlines the dataset, models architecture, and training setup, which form the basis for the analyses that follow. All code and data used in this study are available for reproducibility at our public repository.\footnote{\url{https://github.com/iug-htw/GPTAndPrejudice/}}

\subsection{Dataset}
\label{sec:dataset}

The dataset for this study consists of 37 novels by ten female authors from the late eighteenth and nineteenth centuries: Jane Austen, Anne Brontë, Charlotte Brontë, Emily Brontë, Elizabeth Gaskell, Fanny Burney, George Eliot, Maria Edgeworth, Mary Shelley, and Susan Ferrier. A full list of included works is provided in Appendix~\ref{appendix:full_dataset_authors}. All novels were sourced from Project Gutenberg \cite{projectgutenberg} and were cleaned by removing paratext such as prefaces and footnotes, standardizing chapter divisions, and normalizing spelling, punctuation, and encoding. The final dataset contains about \textbf{7.6 million tokens}, which were tokenized using the GPT-2 tokenizer (tiktoken) with a vocabulary size of 50,257.

These authors were chosen for the thematic coherence of their works. Across their novels we find recurring concerns with gender roles, marriage, class, morality, and individual agency. For example, Jane Austen is noted for her sharp social commentary and portrayals of women’s limited options in aristocratic society \cite{phillips2022jane}; Elizabeth Gaskell engages with class and industrialization while reshaping gender roles \cite{Algotsson2015}; and the Brontës often depict moral struggle and the tension between desire and social constraint \cite{bronte_jane_eyre, bronte_wuthering_heights}.

Because such themes (\textit{gender}, \textit{marriage}, \textit{class}, \textit{emotions}, and \textit{duty}) are shared across the corpus, it provides a coherent and historically grounded setting for examining how language models encode not only literary style but also the conceptual structures of social and moral life. While limited compared to commercial LLM datasets, its small, contained scale reduces noise and variability, allowing for more precise analysis of how models encode and reproduce social constructs.

\subsection{Language Model Design}
\label{sec:lm_design}

The custom model used in this study is a minimalist transformer-based generative language model designed for next-token prediction. The architecture closely follows the implementation detailed in Raschka’s \textit{Build a Large Language Model (From Scratch)} \cite{Raschka2024BuildLLM}. The trained model is publicly available on Hugging Face.\footnote{\url{https://huggingface.co/HTW-KI-Werkstatt/gpt_and_prejudice/}}

\subsubsection{Model Architecture}
The model follows a standard decoder-only Transformer design with an \textbf{896-dimensional} token embedding layer and learned positional embeddings. Dropout (rate 0.2) is applied to enhance generalization. The core of the model consists of \textbf{8 Transformer} Blocks. Each Transformer block first normalizes the input, then applies multi-head (\textbf{14 heads}) self-attention to capture token relationships, followed by dropout and a residual connection. The output is normalized again, then passed through a feedforward network, followed by dropout and another residual connection. 
After all layers, a final normalization is applied, and the hidden states are projected to vocabulary logits for next-token prediction. The model is trained only for next token generation and is \textit{not} instruction fine-tuned. For interpretability, the model records hidden states and attention weights at each layer for later analysis.

\subsubsection{Training Setup}
The AdamW optimizer was used during training with a learning rate of 3e-4 and weight decay of 3e-2. The training loss smoothly declined but plateaued at ~3.4, with a validation loss of ~3.8. While suboptimal for larger neural networks, this was deemed satisfactory given the dataset constraints.

\subsubsection{Model Evaluation}
The model was evaluated using perplexity, a standard measure of predictive accuracy. On the held-out validation set it achieved a perplexity of \textbf{60.4}, showing strong fit to its training domain. To test generalization, we evaluated on three novels by female authors contemporary to those in the training set—\textit{Harriet Martineau}, \textit{Julia Kavanagh}, and \textit{Mary Brunton}—where perplexity rose to \textbf{72.4}.

For context, perplexity values in prior work range from 78.4 on Penn Tree Bank with LSTMs \cite{grave2016improvingneurallanguagemodels} to 29.41 on WikiText-2 with GPT-2 Small (117M parameters) \cite{Radford2019LanguageMA}, with GPT-2 also reporting 99.3 on WikiText-2. These comparisons are not direct baselines, but they help situate the scale of our results.

\begin{table}[H]
\caption{Example of model-generated text reflecting corpus style. Original prompt highlighted in bold.}
\label{tab:example_generations}
\vspace{2ex}
\centering
\begin{tabular}{p{8cm}}
\toprule
\textbf{Generated Text Sample:} \\
\vspace{0.1em}
\textit{\textbf{"you must }}not go to church, you know," said she, with a faint smile, "and I hope you will not be able to stay with me."

"I am very sorry," answered she \\
\vspace{0.1em}
\textit{\textbf{Analysis:}} \textit{Captures period-appropriate dialogue and tone; The exchange lacks clear logical coherence.} \\
\bottomrule
\end{tabular}
\end{table}

We further assessed the model through qualitative inspection of the model’s text generation. The outputs reproduce the sentence structures, word choices, and ironic tone characteristic of nineteenth-century women’s writing, producing grammatically correct and contextually plausible sentences. However, coherence weakens in longer passages as narratives drift, logical consistency fades, and contradictions appear. As shown in Table \ref{tab:example_generations}, the generated text often sounds like it belongs in a nineteenth-century novel but lacks the semantic depth and sustained coherence expected by a human reader.

Since the aim of this study is to examine how LLMs reflect knowledge from their training data rather than to build a fully usable language model, this lack of coherence is not a limitation for our analysis.

\subsection{Sparse Autoencoders Design}
\label{sec:sae_design}

\subsubsection{SAE architecture}
Eight sparse autoencoders were trained, one per transformer layer. Each SAE mirrored the model’s hidden state size (896 neurons) in its input and output layers, with the hidden dimension expanded according to depth: a factor of 3 for Layers 1-2 (2688), 4 for Layers 3-5 (3584), and 5 for Layers 6-8 (4480). This depth-aware scaling reflects the increasing feature density across layers, where early layers capture lexical or syntactic patterns and deeper layers encode more abstract, entangled semantics \cite{jawahar-etal-2019-bert}. Providing greater capacity at depth allows the SAEs to better disentangle overlapping concepts without overparameterizing shallow layers.

Sparsity was enforced using a \textbf{top-$k$ activation} function, retaining only the 50 most active hidden units per sample while suppressing the rest \cite{makhzani2014ksparseautoencoders}. This constraint directs the SAE toward extracting informative features from the model’s hidden states. Training used a simple reconstruction objective with Mean Squared Error (MSE), ensuring focus on accurately reproducing the original activations.

\subsubsection{Training Setup for the SAEs}
All SAEs were trained using the same dataset as the LLM. The text was first split into sentences, then filtered to remove extreme cases (sentences shorter than five words or longer than sixty). Each sentence was fed to the trained LLM, and the hidden states from all layers were extracted from the model and saved in separate files per layer. These embeddings formed the training datasets for the SAE, with a 90:10 train-validation split.

Training ran for up to 500 epochs with early stopping if no improvement was seen after 10 consecutive epochs. The objective throughout was to minimize reconstruction loss while maintaining sparsity in the activations.

\subsection{SAEs Evaluation}
\label{sec:3_4}

To assess the quality of our SAEs, we report reconstruction MSE and cosine similarity across layers (Table \ref{tab:sae_metrics}). While MSE increases with depth (0.13 at Layer 1 to 1.67 at Layer 8), cosine similarity remains consistently high, with mid-layers peaking around 0.89. These values are in line with prior SAE studies. \cite{cunningham2023} observed that reconstruction fidelity declines in deeper layers even as directional structure is preserved, while \cite{lieberum-etal-2024-gemma} showed that fidelity varies across layers but remains sufficiently high for probing tasks. Nonetheless, cosine similarity stays above 0.80 throughout, indicating that the SAEs preserve the directional structure of activations even when magnitude reconstruction weakens. This provides strong evidence that our SAEs are well-calibrated for probing conceptual representations in the model.

\begin{table}[t]
\caption{Layer-wise reconstruction metrics for the eight sparse autoencoders, reporting mean squared error (MSE) and cosine similarity for each layer.}
\label{tab:sae_metrics}
\vspace{2ex}
\footnotesize
\centering
\renewcommand{\arraystretch}{1.3} 
\setlength{\tabcolsep}{10pt}
\begin{tabularx}{\columnwidth}{lXX}
\toprule
\textbf{Layer} & \textbf{Reconstruction MSE} & \textbf{Cosine Similarity} \\
\midrule
1 & 0.128 & 0.705 \\
2 & 0.201 & 0.852 \\
3 & 0.356 & 0.877 \\
4 & 0.505 & 0.897 \\
5 & 0.753 & 0.878 \\
6 & 1.077 & 0.855 \\
7 & 1.385 & 0.830 \\
8 & 1.673 & 0.828 \\
\bottomrule
\end{tabularx}
\end{table}




\section{NEURON AUDITING SETUP}
To help with our analysis, we constructed a probing dataset of short prompts designed to reliably activate targeted themes. Using SAEs and the probing dataset, we trace how different social and narrative themes are reflected in the model’s latent space. An overview of the pipeline is provided in Figure~\ref{fig:pipeline_illust}.

The dataset is fully annotated with multi-hot labels corresponding to 11 recurrent concepts in the novels: \textit{gender (female)}, \textit{gender (male)}, \textit{family}, \textit{marriage}, \textit{wealth}, \textit{emotion}, \textit{love}, \textit{scandal}, \textit{duty}, \textit{class}, and \textit{society}. Each prompt may evoke one or multiple concepts simultaneously. For example, "The girl" is annotated as female, while "His wife" activates marriage, female gender, and male gender.

This annotation scheme reflects the multi-conceptual nature of natural language, where a single phrase often encodes overlapping social and moral dimensions. By framing the dataset as multi-label rather than single-label, we ensure closer alignment with the way such constructs co-occur in the novels themselves.

The final dataset contains \textbf{665 curated prompts}, balanced across the 11 concepts. This controlled resource provides a test bed for mapping SAE neurons to human-interpretable categories, enabling us to track how conceptual structure evolves across the eight layers of the model.

\begin{table*}[t]
\centering
\small
\setlength{\tabcolsep}{3pt}
\caption{Summary of neuron-concept mapping results. (A) Layer-wise statistics, showing the number of selective neurons, growth relative to the previous layer, mean primary AP, and mean polarity for each of the eight layers. (B) Aggregated concept-level statistics across all layers, reporting the total number of primary neurons, mean primary AP, mean polarity, and number of neurons without a secondary concept for the most frequent concepts.}
\label{tab:results_compact_overview}
\vspace{2ex}

\begin{minipage}[t]{0.48\linewidth}
\centering
\textbf{(A) Layer-wise trends}
\setlength{\tabcolsep}{9pt}
\begin{tabularx}{\columnwidth}{lllll}
\toprule
\textbf{L} & \textbf{Selective} & \textbf{Growth} & \textbf{Mean AP} & \textbf{Mean Polarity} \\
\midrule
1 &   4 &   0 & 0.223 & 0.127 \\
2 &   9 &   5 & 0.265 & 0.407 \\
3 &  20 &  11 & 0.244 & 0.328 \\
4 &  34 &  14 & 0.247 & 0.266 \\
5 &  62 &  28 & 0.238 & 0.268 \\
6 & 105 &  43 & 0.228 & 0.239 \\
7 & 147 &  42 & 0.223 & 0.220 \\
8 & 142 &  -5 & 0.222 & 0.230 \\
\bottomrule
\end{tabularx}
\end{minipage}\hfill
\begin{minipage}[t]{0.49\linewidth}
\centering
\textbf{(B) Top concepts (for \S\ref{sec:concept_strengths})}
\renewcommand{\arraystretch}{1.14}
\begin{tabularx}{\columnwidth}{l p{1cm} p{2cm} p{1.6cm} p{1.5cm}}
\toprule
\textbf{Concept} & \textbf{\#} & \textbf{Mean \newline Primary AP} & \textbf{Mean \newline Polarity} & \textbf{No \newline Secondary} \\
\midrule
male    &  74 & 0.309 & 0.302 &  6 \\
society &  77 & 0.309 & 0.293 &  2 \\
female  & 121 & 0.304 & 0.265 & 10 \\
marriage&  80 & 0.274 & 0.283 &  7 \\
family  &  77 & 0.274 & 0.138 &  -- \\
wealth  &  22 & 0.216 & 0.112 &  -- \\
\bottomrule
\end{tabularx}

\end{minipage}
\vspace{2ex}
\end{table*}

\section{CONCEPTS ASSOCIATION MAPPING}
\label{sec:5_mapping}

To link sparse autoencoder activations with the 11 concepts in the probing dataset, we designed a multi-step pipeline. Each probing prompt was passed through the GPT model, and the hidden states from each layer were encoded and reconstructed by the corresponding SAE. For each SAE, we collected neuron activations across the full dataset and applied a selectivity filter: a neuron was retained if its activation exceeded 5.0 and it fired in at least 5 and at most 150 prompts, ensuring it was neither trivially inactive nor overly generic.

For each neuron-concept pair, we computed the following metrics:

\begin{itemize}
    \item \textbf{Average Precision (AP):} How well the neuron ranks prompts containing the concept above those without it (high AP = strong detector).
    \item \textbf{P($\text{fire}\mid$ 1):} Probability that the neuron fires given that the prompt has the concept.
    \item \textbf{P($\text{fire}\mid$ 0):} Probability that the neuron fires given that the prompt does not have the concept.
\end{itemize}

From these, we calculated the difference in firing rates:
\[
\Delta P = P(\text{fire} \mid \text{label}=1) \;-\; P(\text{fire} \mid \text{label}=0)
\]
A positive $\Delta P$ indicates that the neuron fires more often in the presence of the concept, while $\Delta P \leq 0$ implies no useful selectivity or even negative correlation. We therefore restricted analysis to neurons with $\Delta P > 0$, treating them as genuine detectors.


Detector neurons were then ranked by AP. For each neuron, the top-ranked concept was assigned as the \textbf{primary} association, and the second-ranked as a \textbf{secondary} association if sufficiently strong. To compare the relative strength of the two, we computed a polarity score:

\[
\text{Polarity} = \frac{AP_\text{primary} - AP_\text{secondary}}{AP_\text{primary} + 10^{-9}}
\]

This measure expresses how much stronger the primary association is relative to the secondary, normalized by the primary score. A small constant is added to the denominator ($\epsilon = 10^{-9}$) to avoid division by zero in cases where the primary AP is extremely small. Neurons were then categorized as \textit{dominant} (secondary AP $\leq$ 80\% of primary AP), \textit{two-strong} ($\leq$ 50\% margin), or \textit{leaning} ($\leq$ 20\% margin).

The result is a neuron-concept dictionary linking each selective neuron to at most two positively correlated concepts. This dictionary provides the foundation for the analyses presented in the following sections.


\section{RESULTS} 

A summary of the results of mapping sparse features to the concepts present in the probing dataset is given in Table~\ref{tab:results_compact_overview}. Unless otherwise stated, the results in this subsection are aggregated across all eight layers. 

\subsection{General Trends}

As shown in Table \ref{tab:results_compact_overview}A, the number of selective neurons increased steadily with depth, from only 4 in Layer 1 to a peak of 147 in Layer 7, before dropping slightly in Layer 8 (142). This reflects the greater density of features encoded in later layers of the model.

Precision, measured by mean AP across all neuron-concept pairs, varies only modestly across the model. Scores begin at 0.223 in Layer~1, rise to a peak of 0.265 in Layer~2, and then stabilize in the 0.23 range for the remaining layers. This slight peak suggests that early-mid layers yield somewhat more reliable detectors, while deeper layers converge toward broader, less distinct precision. Still, the low mean scores mask the presence of individual neurons with far stronger precision. As shown in Table~\ref{tab:top_detectors}, several units achieve AP values above 0.55 demonstrating that even in a modest model, sharply tuned concept detectors emerge alongside the broader distribution of weaker signals.

Polarity measures follow a similar trajectory. While some neurons in earlier layers were highly selective (mean polarity = 41\% at Layer 2), average polarity scores decreased as depth increased, stabilizing around 22\%-23\% in Layers 7-8. Similarly, the AP gap between primary and secondary concepts shrank across depth (from 0.135 in L2 to 0.068 in L8), indicating that later neurons are increasingly polysemantic, combining multiple concepts rather than responding to a single one.

These trends reveal a clear developmental pattern: early layers contain few but relatively pure detectors, mid-layers strike a balance between clarity and capacity, and deeper layers host the largest number of selective neurons, though these tend to be more entangled across concepts.

\vspace{1ex}
\begin{table}[H]
\caption{Top detector neurons (sorted by primary AP)}
\label{tab:top_detectors}
\vspace{2ex}
\centering
\setlength{\tabcolsep}{8pt}
\begin{tabularx}{\columnwidth}{llllll}
\toprule
\textbf{ID} & \textbf{Layer} & \textbf{Primary} & \textbf{AP} & \textbf{Secondary} & \textbf{Polarity}\\
\midrule    
    4328 & 6 & male   & 0.74 &   --   & 1.00 \\
    1773 & 7 & male   & 0.67 & family & 0.63 \\
    2246 & 5 & male   & 0.64 & family & 0.67 \\
    2130 & 4 & male   & 0.61 & family & 0.63 \\
    2130 & 3 & male   & 0.59 & love   & 0.75 \\
    1466 & 2 & male   & 0.56 &   --   & 1.00 \\
    121  & 7 & family & 0.56 & female & 0.47 \\
\bottomrule
\end{tabularx}
\end{table}

\subsection{Concept Strengths}
\label{sec:concept_strengths}

Not all social and moral categories are represented equally in the model. Some emerge as clear, high-confidence detectors, while others appear mainly in entangled or secondary form (Table~\ref{tab:results_compact_overview}B).

The strongest detectors correspond to gendered categories with mean AP values above 0.30, high polarity scores, and several units acting as near-monosemantic detectors (Table \ref{tab:top_detectors}). For example, one neuron in Layer 6 detects male with an AP of 0.74 and no secondary association, making it the clearest single detector in the sparse space. Alongside gender, \textit{society} also emerges as a relatively strong theme, suggesting that broad social structures are represented with more coherence than private or affective notions.

\textit{Marriage} and \textit{family} are also frequent as primary associations (80 and 77 neurons respectively), but they tend to be less pure. Their mean primary AP scores are lower ($\sim$0.27), and their mean polarity values (\textit{family} = 14\%, \textit{marriage} = 28\%) show that these neurons are more often entangled with related themes such as \textit{wealth}, \textit{female}, or \textit{male}.

Other categories, such as \textit{emotion}, \textit{duty}, and \textit{love}, appear in smaller numbers and with weaker selectivity. \textit{Wealth} rarely appears as a primary concept but is the fourth most common secondary association, while \textit{scandal} barely registers (1 neuron). These categories appear less as standalone detectors and more as modifiers embedded within other conceptual contexts.

These trends suggest that the model \textbf{most strongly encodes gendered categories (male/female) and broad social structures (society)} as coherent detectors, while themes like \textbf{marriage and family are richly represented but more polysemantic}. Economic and moral dimensions such as wealth or duty are present but usually in secondary or entangled form, rather than as dominant signals.

\begin{figure*}[t]
    \centering
    \includegraphics[width=\textwidth]{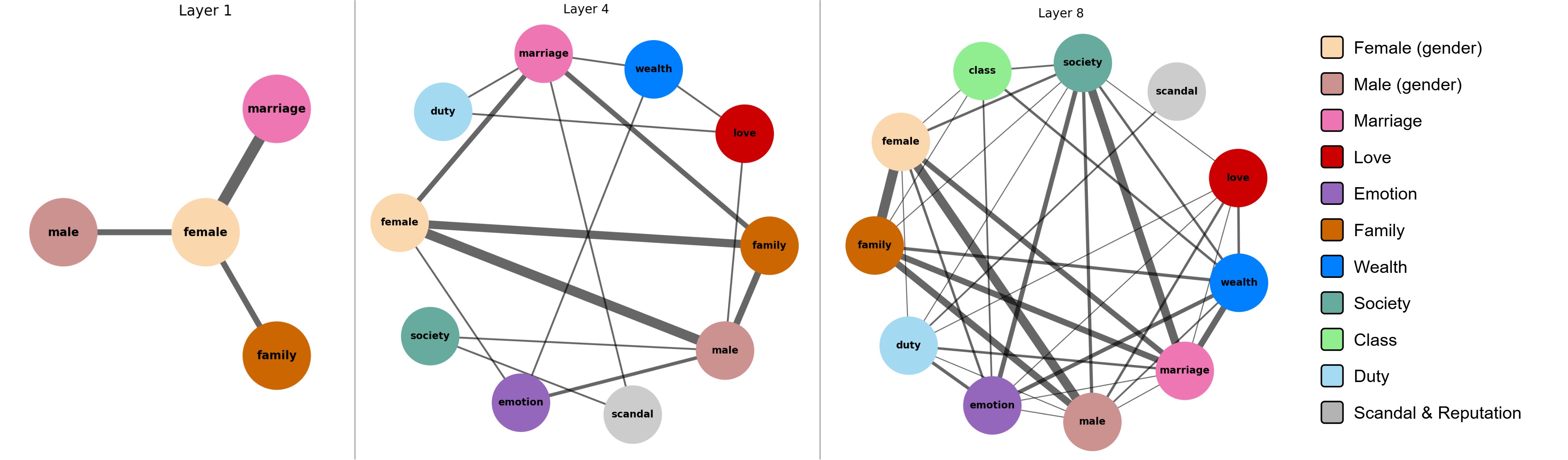}
    \caption{Conceptual mappings of dual-theme neurons in SAE Layer 1 , 4, and 8 (from left to right). Each node represents a social concept, and edges represent shared neurons that consistently co-activate for both themes. }
    \label{fig:conecepts-graph}
    \vspace{3ex}
\end{figure*}

\subsection{Dual-Themed Neurons}
Dual-themed neurons, which are units with both a primary and a secondary concept, show how social and moral categories overlap in the model. They expose relationships rather than single detectors. 
Figure~\ref{fig:conecepts-graph} presents three representative layers (Layers 1, 4, and 8) corresponding to early, mid, and late stages of the model. These layers serve as case studies of conceptual evolution. In each graph, nodes denote concepts and edges indicate the number of neurons shared between them, with edge width proportional to the frequency of co-activation normalized per layer. The full set of graphs for all eight layers is provided in Appendix~\ref{appendix:all_layers_graphs}.

\textbf{In Layer 1}, the network is extremely sparse, with only a few selective neurons forming a simple backbone around \textit{female}, making it the central node at this depth. This suggests that gender (especially \textit{female}) anchors the earliest structure, with kinship and marriage clustered immediately around it.

\textbf{By Layer 4}, the network expands to 16 unique pairings, with a clear cluster around gender and kinship. \textit{Female}, \textit{male}, \textit{family}, and \textit{marriage} dominate as central nodes, connected by multiple overlapping neurons. At the same time, we see the first outward ties to broader social categories: \textit{wealth}, \textit{scandal}, and \textit{society}. Emotion also enters here in tentative form (\textit{emotion–male}, \textit{emotion–female}), layering affective tones onto the gendered core.

\textbf{In Layer 8}, the network is densely interconnected. The \textit{gender–family–marriage} cluster remains, but \textit{marriage} becomes the key bridge outward, linking strongly to \textit{society} and \textit{wealth}. Emotional and moral themes also enter, with edges such as \textit{duty–emotion} and \textit{duty–marriage}. Notably, \textit{love–marriage} remains weak, and \textit{scandal} and \textit{class} remain peripheral.

Across all layers, the \textit{gender–family–marriage} triad persists as a stable backbone, growing more interconnected with depth even as its dominance gives way to broader entanglement. Together, these three layers illustrate the progression of dual-themed associations: sparse and focused in the early model, more coherent and structured at mid-depth, and densely interwoven in later layers, where boundaries blur but recurring sub-clusters remain.


\section{DISCUSSION}

The main goal of this study is not to interpret a model in isolation but to demonstrate how sparse autoencoders combined with a probing dataset can recover conceptual relations from text. Using a GPT-style model trained on nineteenth-century novels, we identified recurring structures around social and moral themes from the underlying corpus. This serves as a proof of concept for a pipeline that can be scaled to larger models and corpora, offering a generalizable approach to studying cultural and social patterns in text.

The associations we uncover align with themes well documented in literary scholarship. For instance, a persistent \textbf{gender-family-marriage} backbone reflects how central such themes are to the novels. For example, in George Eliot’s \textit{Middlemarch}, Brooke’s marriage illustrates the tension between gender expectations, duty, and family standing, while in Anne Brontë’s \textit{The Tenant of Wildfell Hall}, Helen’s abusive marriage shows how familial obligations confine women in destructive unions \cite{Senf1990TheTO, kelly2010middlemarch}.

The weak \textbf{marriage–love} link likewise matches the period’s cultural logic, where marriage was framed as rational and stabilizing, while love was considered unstable and rarely central \cite{kelly2010middlemarch}. \textit{Middlemarch} itself depicts love as disconnected from a “good match.” By contrast, the model highlights a strong \textbf{marriage–wealth–society} axis, consistent with the economic framing of marriage. In Austen’s \textit{Pride and Prejudice}, a character defends her acceptance of a marriage proposal for financial security, while in Charlotte Brontë’s \textit{Jane Eyre}, Jane’s inheritance grants independence and enables a more equal union \cite{Sudesh2022,bronte_jane_eyre}. These examples confirm that the model’s relational framework mirrors the thematic structures in the corpus.

Another notable pattern emerges. The \textit{male} category produces the purest detectors, with several neurons highly selective for \textit{male} alone. \textit{Female}, by contrast, rarely appears in isolation, with only a few pure detectors. In the dual-themed networks, however, \textit{female} takes on a central role, outwardly (with \textit{female} as primary) linking to nearly every major concept, while \textit{male} forms fewer, more selective ties, often tied back to \textit{female}. This suggests that \textit{male} is encoded more discretely, while \textit{female} functions as a relational hub. One possible explanation, though speculative, is the narrative perspective. These novels are written from a female viewpoint, often with \textit{male} figures as objects of description.

It is important to recognize that these results are shaped by the probing dataset. We defined eleven concepts that guided the analysis, and the relations surfaced are therefore dependent on this design. Other categories (e.g., \textit{religion}, \textit{education}, or \textit{politics}) might have yielded different insights from the data. Some, like \textit{society}, \textit{class} or \textit{scandal}, were harder to capture in short probing prompts, reflecting both the broadness of their definitions and the subjectivity of human annotation. Rather than a flaw, we see this as a feature of the method: the probing dataset can be tuned to the needs of a given application.

The pipeline extends beyond literary analysis, since the probing dataset can be tailored to different domains. In HR, categories such as gender, occupation, promotion, and salary could be used to evaluate bias patterns, while in healthcare, probing might focus on gender, diagnosis, treatment, and insurance coverage. This adaptability makes the method a customizable lens for examining model behavior across contexts.
While demonstrated here on a small model and a limited dataset, the ability to recover coherent thematic structures in this setting demonstrates the viability of the approach. Applied to larger models trained on more heterogeneous corpora, the same pipeline could uncover richer and more nuanced patterns.

Most importantly, the contribution here goes beyond interpretability. The LLMs+SAEs framework uses models as instruments to study corpora themselves. Rather than treating SAEs only as tools for explaining models, this approach surfaces the social patterns and thematic structures embedded in training data. It thus provides a scalable framework for exploring literary traditions at scale and auditing massive datasets that cannot be examined manually, bridging the gap between interpretability research and corpus analysis.


\section{CONCLUSION}

This paper presented a pipeline that combines LLMs with SAEs to uncover conceptual relations in text. A GPT-style model was first trained on a corpus of nineteenth-century novels, and SAEs were then applied to its hidden states. The resulting sparse features were mapped to 11 social and moral categories, revealing recurring structures that reflect the literature. 

The results show that the LLM+SAEs approach can offer a systematic method for exploring cultural and social patterns in corpora. Because the probing dataset can be customized, the same pipeline can be applied to domains ranging from literary analysis to fairness auditing in applied systems.

Future work includes applying the pipeline to larger, more diverse corpora and extending the probing concepts to test different conceptual frames.


\begin{ack}
This paper portrays the work carried out in the context of the KIWI project (16DHBKI071) that is generously funded by the Federal Ministry of Research, Technology and Space (BMFTR).
\end{ack}


\bibliography{mybibfile}

\begin{thebibliography}{40}
\providecommand{\natexlab}[1]{#1}
\providecommand{\url}[1]{\texttt{#1}}
\expandafter\ifx\csname urlstyle\endcsname\relax
  \providecommand{\doi}[1]{doi: #1}\else
  \providecommand{\doi}{doi: \begingroup \urlstyle{rm}\Url}\fi

\bibitem[Algotsson(2015)]{Algotsson2015}
A.~Algotsson.
\newblock {Transgression and Tradition: Redefining Gender Roles in Elizabeth Gaskell´s North and South (Dissertation)}, 2015.
\newblock URL \url{{https://urn.kb.se/resolve?urn=urn:nbn:se:liu:diva-119026}}.
\newblock \url{https://urn.kb.se/resolve?urn=urn:nbn:se:liu:diva-119026}.

\bibitem[Baek et~al.(2024)Baek, Li, and Tegmark]{baek2024generalizationstarvationhintsuniversality}
D.~D. Baek, Y.~Li, and M.~Tegmark.
\newblock {Generalization from Starvation: Hints of Universality in LLM Knowledge Graph Learning}, 2024.
\newblock URL \url{https://arxiv.org/abs/2410.08255}.
\newblock \url{https://arxiv.org/abs/2410.08255}.

\bibitem[Bender et~al.(2021)Bender, Gebru, McMillan-Major, and Shmitchell]{10.1145/3442188.3445922}
E.~M. Bender, T.~Gebru, A.~McMillan-Major, and S.~Shmitchell.
\newblock {On the Dangers of Stochastic Parrots: Can Language Models Be Too Big?}
\newblock In \emph{Proceedings of the 2021 ACM Conference on Fairness, Accountability, and Transparency}, FAccT '21, New York, NY, USA, 2021. Association for Computing Machinery.
\newblock ISBN 9781450383097.
\newblock \doi{10.1145/3442188.3445922}.
\newblock URL \url{https://doi.org/10.1145/3442188.3445922}.

\bibitem[Bricken et~al.(2023)Bricken, Templeton, Batson, Chen, Jermyn, Conerly, Turner, Anil, Denison, Askell, Lasenby, Wu, Kravec, Schiefer, Maxwell, Joseph, Hatfield-Dodds, Tamkin, Nguyen, McLean, Burke, Hume, Carter, Henighan, and Olah]{bricken2023monosemanticity}
T.~Bricken, A.~Templeton, J.~Batson, B.~Chen, A.~Jermyn, T.~Conerly, N.~Turner, C.~Anil, C.~Denison, A.~Askell, R.~Lasenby, Y.~Wu, S.~Kravec, N.~Schiefer, T.~Maxwell, N.~Joseph, Z.~Hatfield-Dodds, A.~Tamkin, K.~Nguyen, B.~McLean, J.~E. Burke, T.~Hume, S.~Carter, T.~Henighan, and C.~Olah.
\newblock {Towards Monosemanticity: Decomposing Language Models With Dictionary Learning}.
\newblock \emph{Transformer Circuits Thread}, 2023.
\newblock \url{https://transformer-circuits.pub/2023/monosemantic-features/index.html}.

\bibitem[Brontë(1998)]{bronte_jane_eyre}
C.~Brontë.
\newblock \emph{{Jane Eyre: An Autobiography}}.
\newblock Project Gutenberg, 1998.
\newblock URL \url{https://www.gutenberg.org/ebooks/1260}.
\newblock EBook \#1260. \url{https://www.gutenberg.org/ebooks/1260}. Accessed: 2025-09-16.

\bibitem[Brontë(1996)]{bronte_wuthering_heights}
E.~Brontë.
\newblock \emph{{Wuthering Heights}}.
\newblock Project Gutenberg, 1996.
\newblock URL \url{https://www.gutenberg.org/ebooks/768}.
\newblock EBook \#768. \url{https://www.gutenberg.org/ebooks/768}. Accessed: 2025-09-16.

\bibitem[Brown et~al.(2020)Brown, Mann, Ryder, Subbiah, Kaplan, Dhariwal, Neelakantan, Shyam, Sastry, Askell, Agarwal, Herbert-Voss, Krueger, Henighan, Child, Ramesh, Ziegler, Wu, Winter, Hesse, Chen, Sigler, Litwin, Gray, Chess, Clark, Berner, McCandlish, Radford, Sutskever, and Amodei]{brown2020languagemodelsfewshotlearners}
T.~B. Brown, B.~Mann, N.~Ryder, M.~Subbiah, J.~Kaplan, P.~Dhariwal, A.~Neelakantan, P.~Shyam, G.~Sastry, A.~Askell, S.~Agarwal, A.~Herbert-Voss, G.~Krueger, T.~Henighan, R.~Child, A.~Ramesh, D.~M. Ziegler, J.~Wu, C.~Winter, C.~Hesse, M.~Chen, E.~Sigler, M.~Litwin, S.~Gray, B.~Chess, J.~Clark, C.~Berner, S.~McCandlish, A.~Radford, I.~Sutskever, and D.~Amodei.
\newblock Language models are few-shot learners.
\newblock In \emph{Proceedings of the 34th International Conference on Neural Information Processing Systems}, NIPS '20, Red Hook, NY, USA, 2020. Curran Associates Inc.
\newblock ISBN 9781713829546.

\bibitem[Chughtai et~al.(2023)Chughtai, Chan, and Nanda]{chughtai2023toymodeluniversalityreverse}
B.~Chughtai, L.~Chan, and N.~Nanda.
\newblock {A toy model of universality: reverse engineering how networks learn group operations}.
\newblock In \emph{Proceedings of the 40th International Conference on Machine Learning}, ICML'23. JMLR.org, 2023.

\bibitem[Dodge et~al.(2021)Dodge, Sap, Marasovi{\'c}, Agnew, Ilharco, Groeneveld, Mitchell, and Gardner]{dodge2021documentinglargewebtextcorpora}
J.~Dodge, M.~Sap, A.~Marasovi{\'c}, W.~Agnew, G.~Ilharco, D.~Groeneveld, M.~Mitchell, and M.~Gardner.
\newblock {Documenting Large Webtext Corpora: A Case Study on the Colossal Clean Crawled Corpus}.
\newblock In \emph{Proceedings of the 2021 Conference on Empirical Methods in Natural Language Processing}, Online and Punta Cana, Dominican Republic, Nov. 2021. Association for Computational Linguistics.
\newblock \doi{10.18653/v1/2021.emnlp-main.98}.
\newblock URL \url{https://aclanthology.org/2021.emnlp-main.98/}.

\bibitem[Dong et~al.(2024)Dong, Wang, Yu, and Caverlee]{dong2024disclosuremitigationgenderbias}
X.~Dong, Y.~Wang, P.~S. Yu, and J.~Caverlee.
\newblock {Disclosure and Mitigation of Gender Bias in LLMs}, 2024.
\newblock URL \url{https://arxiv.org/abs/2402.11190}.
\newblock \url{https://arxiv.org/abs/2402.11190}.

\bibitem[Elhage et~al.(2022)Elhage, Hume, Olsson, Schiefer, Henighan, Kravec, Hatfield-Dodds, Lasenby, Drain, Chen, Grosse, McCandlish, Kaplan, Amodei, Wattenberg, and Olah]{elhage2022toymodelssuperposition}
N.~Elhage, T.~Hume, C.~Olsson, N.~Schiefer, T.~Henighan, S.~Kravec, Z.~Hatfield-Dodds, R.~Lasenby, D.~Drain, C.~Chen, R.~Grosse, S.~McCandlish, J.~Kaplan, D.~Amodei, M.~Wattenberg, and C.~Olah.
\newblock {Toy Models of Superposition}, 2022.
\newblock URL \url{https://arxiv.org/abs/2209.10652}.
\newblock {https://arxiv.org/abs/2209.10652}.

\bibitem[Gallegos et~al.(2024)Gallegos, Rossi, Barrow, Tanjim, Kim, Dernoncourt, Yu, Zhang, and Ahmed]{gallegos2024biasfairnesslargelanguage}
I.~O. Gallegos, R.~A. Rossi, J.~Barrow, M.~M. Tanjim, S.~Kim, F.~Dernoncourt, T.~Yu, R.~Zhang, and N.~K. Ahmed.
\newblock {Bias and Fairness in Large Language Models: A Survey}.
\newblock \emph{Computational Linguistics}, 50\penalty0 (3), Sept. 2024.
\newblock \doi{10.1162/coli_a_00524}.
\newblock URL \url{https://aclanthology.org/2024.cl-3.8/}.

\bibitem[Grave et~al.(2017)Grave, Joulin, and Usunier]{grave2016improvingneurallanguagemodels}
E.~Grave, A.~Joulin, and N.~Usunier.
\newblock {Improving Neural Language Models with a Continuous Cache}.
\newblock In \emph{International Conference on Learning Representations}, 2017.
\newblock URL \url{https://openreview.net/forum?id=B184E5qee}.

\bibitem[Hindupur et~al.(2025)Hindupur, Lubana, Fel, and Ba]{hindupur2025projectingassumptionsdualitysparse}
S.~S.~R. Hindupur, E.~S. Lubana, T.~Fel, and D.~E. Ba.
\newblock {Projecting Assumptions: The Duality Between Sparse Autoencoders and Concept Geometry}.
\newblock In \emph{ICML 2025 Workshop on Methods and Opportunities at Small Scale}, 2025.
\newblock URL \url{https://openreview.net/forum?id=AKaoBzhIIF}.

\bibitem[Huben et~al.(2024)Huben, Cunningham, Smith, Ewart, and Sharkey]{cunningham2023}
R.~Huben, H.~Cunningham, L.~R. Smith, A.~Ewart, and L.~Sharkey.
\newblock {Sparse Autoencoders Find Highly Interpretable Features in Language Models}.
\newblock In \emph{The Twelfth International Conference on Learning Representations}, 2024.
\newblock URL \url{https://openreview.net/forum?id=F76bwRSLeK}.

\bibitem[Jawahar et~al.(2019)Jawahar, Sagot, and Seddah]{jawahar-etal-2019-bert}
G.~Jawahar, B.~Sagot, and D.~Seddah.
\newblock What does {BERT} learn about the structure of language?
\newblock In \emph{Proceedings of the 57th Annual Meeting of the Association for Computational Linguistics}, pages 3651--3657, Florence, Italy, July 2019. Association for Computational Linguistics.
\newblock \doi{10.18653/v1/P19-1356}.
\newblock URL \url{https://aclanthology.org/P19-1356/}.

\bibitem[Jing et~al.(2025)Jing, Yao, Ran, Guo, Wang, Hou, and Li]{jing2025sparseautoencoderinterpretslinguistic}
Y.~Jing, Z.~Yao, L.~Ran, H.~Guo, X.~Wang, L.~Hou, and J.~Li.
\newblock {Sparse Auto-Encoder Interprets Linguistic Features in Large Language Models}, 2025.
\newblock URL \url{https://arxiv.org/abs/2502.20344}.
\newblock \url{https://arxiv.org/abs/2502.20344}.

\bibitem[Kelly(2010)]{kelly2010middlemarch}
K.~M. Kelly.
\newblock {George Eliot's Middlemarch: The Making of a Modern Marriage}.
\newblock Master's thesis, University of New Orleans, 2010.
\newblock URL \url{https://scholarworks.uno.edu/td/1173}.
\newblock \url{https://scholarworks.uno.edu/td/1173}.

\bibitem[Kim et~al.(2025)Kim, Li, Spiliopoulou, Ma, Ballesteros, and Wang]{kim2025detectingtrainingdatalarge}
G.~Kim, Y.~Li, E.~Spiliopoulou, J.~Ma, M.~Ballesteros, and W.~Y. Wang.
\newblock {Detecting Training Data of Large Language Models via Expectation Maximization}.
\newblock In \emph{NeurIPS 2025 Workshop on Evaluating the Evolving LLM Lifecycle: Benchmarks, Emergent Abilities, and Scaling}, 2025.
\newblock URL \url{https://openreview.net/forum?id=ZYIP2PhWyz}.

\bibitem[K{\"o}ksal et~al.(2023)K{\"o}ksal, Yalcin, Akbiyik, Kilavuz, Korhonen, and Schuetze]{köksal2023languageagnosticbiasdetectionlanguage}
A.~K{\"o}ksal, O.~Yalcin, A.~Akbiyik, M.~Kilavuz, A.~Korhonen, and H.~Schuetze.
\newblock {Language-Agnostic Bias Detection in Language Models with Bias Probing}.
\newblock In \emph{Findings of the Association for Computational Linguistics: EMNLP 2023}, Singapore, Dec. 2023. Association for Computational Linguistics.
\newblock \doi{10.18653/v1/2023.findings-emnlp.848}.
\newblock URL \url{https://aclanthology.org/2023.findings-emnlp.848/}.

\bibitem[Li et~al.(2021)Li, Lei, Wang, Jiang, and Liu]{LI2021107003}
Y.~Li, Y.~Lei, P.~Wang, M.~Jiang, and Y.~Liu.
\newblock {Embedded stacked group sparse autoencoder ensemble with L1 regularization and manifold reduction}.
\newblock \emph{Applied Soft Computing}, 101:\penalty0 107003, 2021.
\newblock ISSN 1568-4946.
\newblock \doi{https://doi.org/10.1016/j.asoc.2020.107003}.
\newblock URL \url{https://www.sciencedirect.com/science/article/pii/S156849462030942X}.

\bibitem[Li et~al.(2024)Li, Du, Song, Wang, and Wang]{li2024surveyfairnesslargelanguage}
Y.~Li, M.~Du, R.~Song, X.~Wang, and Y.~Wang.
\newblock {A Survey on Fairness in Large Language Models}, 2024.
\newblock URL \url{https://arxiv.org/abs/2308.10149}.
\newblock \url{https://arxiv.org/abs/2308.10149}.

\bibitem[Lieberum et~al.(2024)Lieberum, Rajamanoharan, Conmy, Smith, Sonnerat, Varma, Kramar, Dragan, Shah, and Nanda]{lieberum-etal-2024-gemma}
T.~Lieberum, S.~Rajamanoharan, A.~Conmy, L.~Smith, N.~Sonnerat, V.~Varma, J.~Kramar, A.~Dragan, R.~Shah, and N.~Nanda.
\newblock Gemma scope: Open sparse autoencoders everywhere all at once on gemma 2.
\newblock In \emph{Proceedings of the 7th BlackboxNLP Workshop: Analyzing and Interpreting Neural Networks for NLP}, pages 278--300, Miami, Florida, US, Nov. 2024. Association for Computational Linguistics.
\newblock \doi{10.18653/v1/2024.blackboxnlp-1.19}.
\newblock URL \url{https://aclanthology.org/2024.blackboxnlp-1.19/}.

\bibitem[Ling et~al.(2025)Ling, Rabbi, Wang, and Yang]{Ling_Rabbi_Wang_Yang_2025}
L.~Ling, F.~Rabbi, S.~Wang, and J.~Yang.
\newblock {Bias Unveiled: Investigating Social Bias in LLM-Generated Code}.
\newblock \emph{Proceedings of the AAAI Conference on Artificial Intelligence}, 39\penalty0 (26):\penalty0 27491--27499, Apr. 2025.
\newblock \doi{10.1609/aaai.v39i26.34961}.
\newblock URL \url{https://ojs.aaai.org/index.php/AAAI/article/view/34961}.

\bibitem[Lou et~al.(2025)Lou, Li, Ji, and Yang]{lou2025saevinterpretingmultimodalmodels}
H.~Lou, C.~Li, J.~Ji, and Y.~Yang.
\newblock {SAE-V: Interpreting Multimodal Models for Enhanced Alignment}.
\newblock In \emph{Forty-second International Conference on Machine Learning}, 2025.
\newblock URL \url{https://openreview.net/forum?id=S4HPn5Bo6k}.

\bibitem[Mahran and Simbeck(2025)]{mahran2025multilingualpipeline}
M.~Mahran and K.~Simbeck.
\newblock {Investigating Bias: A Multilingual Pipeline for Generating, Solving, and Evaluating Math Problems with LLMs}.
\newblock In \emph{{Edu4AI'25: 2nd Workshop on Education for Artificial Intelligence, ECAI}}, Bologna, Italy, 2025.
\newblock URL \url{https://arxiv.org/abs/2509.17701}.

\bibitem[Makhzani and Frey(2014)]{makhzani2014ksparseautoencoders}
A.~Makhzani and B.~Frey.
\newblock {k-Sparse Autoencoders}, 2014.
\newblock URL \url{https://arxiv.org/abs/1312.5663}.
\newblock \url{https://arxiv.org/abs/1312.5663}.

\bibitem[Orgad and Belinkov(2022)]{orgad-belinkov-2022-choose}
H.~Orgad and Y.~Belinkov.
\newblock Choose your lenses: Flaws in gender bias evaluation.
\newblock In \emph{Proceedings of the 4th Workshop on Gender Bias in Natural Language Processing (GeBNLP)}, Seattle, Washington, July 2022. Association for Computational Linguistics.
\newblock \doi{10.18653/v1/2022.gebnlp-1.17}.
\newblock URL \url{https://aclanthology.org/2022.gebnlp-1.17/}.

\bibitem[Paulo et~al.(2025)Paulo, Mallen, Juang, and Belrose]{paulo2024automaticallyinterpretingmillionsfeatures}
G.~S. Paulo, A.~T. Mallen, C.~Juang, and N.~Belrose.
\newblock {Automatically Interpreting Millions of Features in Large Language Models}.
\newblock In \emph{Forty-second International Conference on Machine Learning}, 2025.
\newblock URL \url{https://openreview.net/forum?id=EemtbhJOXc}.

\bibitem[Phillips(2022)]{phillips2022jane}
E.~Phillips.
\newblock Jane austen: A study on the influences, world, and character of an eighteenth-century novelist.
\newblock \emph{Bound Away: The Liberty Journal of History}, 5\penalty0 (1), 2022.
\newblock \doi{10.70623/YRIR8982}.
\newblock URL \url{https://digitalcommons.liberty.edu/ljh/vol5/iss1/1}.

\bibitem[{Project Gutenberg}(n.d.)]{projectgutenberg}
{Project Gutenberg}.
\newblock \url{https://www.gutenberg.org/}, n.d.
\newblock Accessed: 2025-09-16.

\bibitem[Radford et~al.(2019)Radford, Wu, Child, Luan, Amodei, and Sutskever]{Radford2019LanguageMA}
A.~Radford, J.~Wu, R.~Child, D.~Luan, D.~Amodei, and I.~Sutskever.
\newblock {Language Models are Unsupervised Multitask Learners}.
\newblock \emph{OpenAI}, 2019.
\newblock URL \url{https://cdn.openai.com/better-language-models/language_models_are_unsupervised_multitask_learners.pdf}.

\bibitem[Rajamanoharan et~al.(2024)Rajamanoharan, Lieberum, Sonnerat, Conmy, Varma, Kramár, and Nanda]{rajamanoharan2024jumpingaheadimprovingreconstruction}
S.~Rajamanoharan, T.~Lieberum, N.~Sonnerat, A.~Conmy, V.~Varma, J.~Kramár, and N.~Nanda.
\newblock {Jumping Ahead: Improving Reconstruction Fidelity with JumpReLU Sparse Autoencoders}, 2024.
\newblock URL \url{https://arxiv.org/abs/2407.14435}.
\newblock \url{https://arxiv.org/abs/2407.14435}.

\bibitem[Raschka(2024)]{Raschka2024BuildLLM}
S.~Raschka.
\newblock \emph{{Build a Large Language Model (From Scratch)}}.
\newblock Manning Publications, Shelter Island, NY, 2024.
\newblock ISBN 978-1633437166.
\newblock URL \url{https://www.manning.com/books/build-a-large-language-model-from-scratch}.

\bibitem[Senf(1990)]{Senf1990TheTO}
C.~A. Senf.
\newblock The tenant of wildfell hall": Narrative silences and questions of gender.
\newblock \emph{College English}, 52:\penalty0 446, 1990.
\newblock URL \url{https://api.semanticscholar.org/CorpusID:151312545}.

\bibitem[Simbeck and Mahran(2025)]{simbeck2025saes}
K.~Simbeck and M.~Mahran.
\newblock {Mechanistic Interpretability with SAEs: Probing Religion, Violence, and Geography in Large Language Models}.
\newblock In \emph{{AEQUITAS 2025: Workshop on Fairness and Bias in AI, ECAI}}, Bologna, Italy, 2025.
\newblock URL \url{https://arxiv.org/abs/2509.17665}.

\bibitem[Simon and Zou(2024)]{Simon2024.11.14.623630}
E.~Simon and J.~Zou.
\newblock {InterPLM: Discovering Interpretable Features in Protein Language Models via Sparse Autoencoders}, 2024.
\newblock URL \url{https://arxiv.org/abs/2412.12101}.

\bibitem[Sudesh(2022)]{Sudesh2022}
Sudesh.
\newblock {Marriage, Love and Money in Themes of Jane Austen’s Mansfield Park and Emma}.
\newblock \emph{Academia Arena}, 14\penalty0 (11):\penalty0 1--7, 2022.
\newblock ISSN 1553-992X.
\newblock URL \url{https://www.sciencepub.net/academia/aaj141122/01_38184aaj141122_1_7.pdf}.

\bibitem[Thaler et~al.(2024)Thaler, Köksal, Leidinger, Korhonen, and Schütze]{thaler2024farbiasgo}
M.~Thaler, A.~Köksal, A.~Leidinger, A.~Korhonen, and H.~Schütze.
\newblock {How far can bias go? -- Tracing bias from pretraining data to alignment}, 2024.
\newblock URL \url{https://arxiv.org/abs/2411.19240}.
\newblock \url{https://arxiv.org/abs/2411.19240}.

\bibitem[Yu et~al.(2025)Yu, Liu, Wu, Chung, Zhang, Li, Yeung, and Zhou]{yu2025stochasticparrotllmsshoulder}
M.~Yu, L.~Liu, J.~Wu, T.~T. Chung, S.~Zhang, J.~Li, D.-Y. Yeung, and J.~Zhou.
\newblock {The Stochastic Parrot on LLM's Shoulder: A Summative Assessment of Physical Concept Understanding}.
\newblock In \emph{Proceedings of the 2025 Conference of the Nations of the Americas Chapter of the Association for Computational Linguistics: Human Language Technologies (Volume 1: Long Papers)}, Albuquerque, New Mexico, Apr. 2025. Association for Computational Linguistics.
\newblock ISBN 979-8-89176-189-6.
\newblock \doi{10.18653/v1/2025.naacl-long.569}.
\newblock URL \url{https://aclanthology.org/2025.naacl-long.569/}.

\end{thebibliography}


\clearpage
\appendix
\onecolumn

\section{FULL LIST OF DATASET AUTHORS AND WORKS}
\label{appendix:full_dataset_authors}

Table~\ref{tab:dataset_authors} lists all authors and novels included in the corpus, along with the general period of their literary activity. All novels were sourced from \textbf{Project Gutenberg} (\url{https://www.gutenberg.org/}).

\begin{table}[H]
\centering
\small
\renewcommand{\arraystretch}{1.2}
\setlength{\tabcolsep}{8pt}
\begin{tabularx}{\linewidth}{lcXl}
\toprule
\textbf{Author} & \textbf{Novels Count} & \textbf{Novels Included} & \textbf{Active Period} \\
\midrule
\multicolumn{3}{l}{\textbf{Training \& Validation Set}} \\
\midrule
Anne Brontë & 2 & \textit{Agnes Grey}; \textit{The Tenant of Wildfell Hall} & 1836--1849 \\
Charlotte Brontë & 4 & \textit{Jane Eyre}; \textit{Shirley}; \textit{Villette}; \textit{The Professor} & 1835--1855 \\
Emily Brontë & 1 & \textit{Wuthering Heights} & 1845--1848 \\
Elizabeth Gaskell & 8 & \textit{Cranford}; \textit{Mary Barton}; \textit{My Lady Ludlow}; \textit{North and South}; \textit{Ruth}; \textit{Sylvia's Lovers}; \textit{The Moorland Cottage}; \textit{Wives and Daughters} & 1848--1865 \\
Frances Burney & 4 & \textit{Evelina}; \textit{Cecilia}; \textit{Camilla}; \textit{The Wanderer} & 1778--1814 \\
George Eliot & 2 & \textit{Middlemarch}; \textit{The Mill on the Floss} & 1857--1880 \\
Jane Austen & 6 & \textit{Pride and Prejudice}; \textit{Sense and Sensibility}; \textit{Emma}; \textit{Mansfield Park}; \textit{Northanger Abbey}; \textit{Persuasion} & 1787--1817 \\
Maria Edgeworth & 8 & \textit{Belinda}; \textit{Castle Rackrent}; \textit{Harrington \& Ormond}; \textit{Helen}; \textit{Leonora}; \textit{Patronage}; \textit{The Absentee}; \textit{Tomorrow} & 1795--1848 \\
Mary Shelley & 1 & \textit{Lodore} & 1817--1851 \\
Susan Ferrier & 1 & \textit{Marriage} & 1810--1831 \\
\midrule
\multicolumn{3}{l}{\textbf{Evaluation Set}} \\
\midrule
Harriet Martineau & 1 & \textit{Deerbrook} & 1823--1876 \\
Julia Kavanagh & 1 & \textit{Daisy Burns} & 1847--1877 \\
Mary Brunton & 1 & \textit{Self-Control} & 1811--1818 \\
\bottomrule
\end{tabularx}
\vspace{2ex}
\caption{Authors and novels included in the training, validation, and evaluation datasets, with approximate active literary periods.}
\label{tab:dataset_authors}
\vspace{5ex}
\end{table}

\section{COMPUTATIONAL ENVIRONMENT}
\label{appendix:infrastructure}

All training was conducted on a high-performance computing (HPC) cluster managed with SLURM. The GPT-style language model and the eight Sparse Autoencoders (SAEs) were trained on nodes equipped with 8× NVIDIA A100 GPUs. Individual training runs did not exceed 2 hours.The code to reproduce, data, and results can be found at \url{https://github.com/BLINDED}.

\newpage

\section{DUAL-THEMED NEURON GRAPHS ACROSS LAYERS}
\label{appendix:all_layers_graphs}

\begin{figure}[H]
    \centering
    \includegraphics[width=0.8\textwidth]{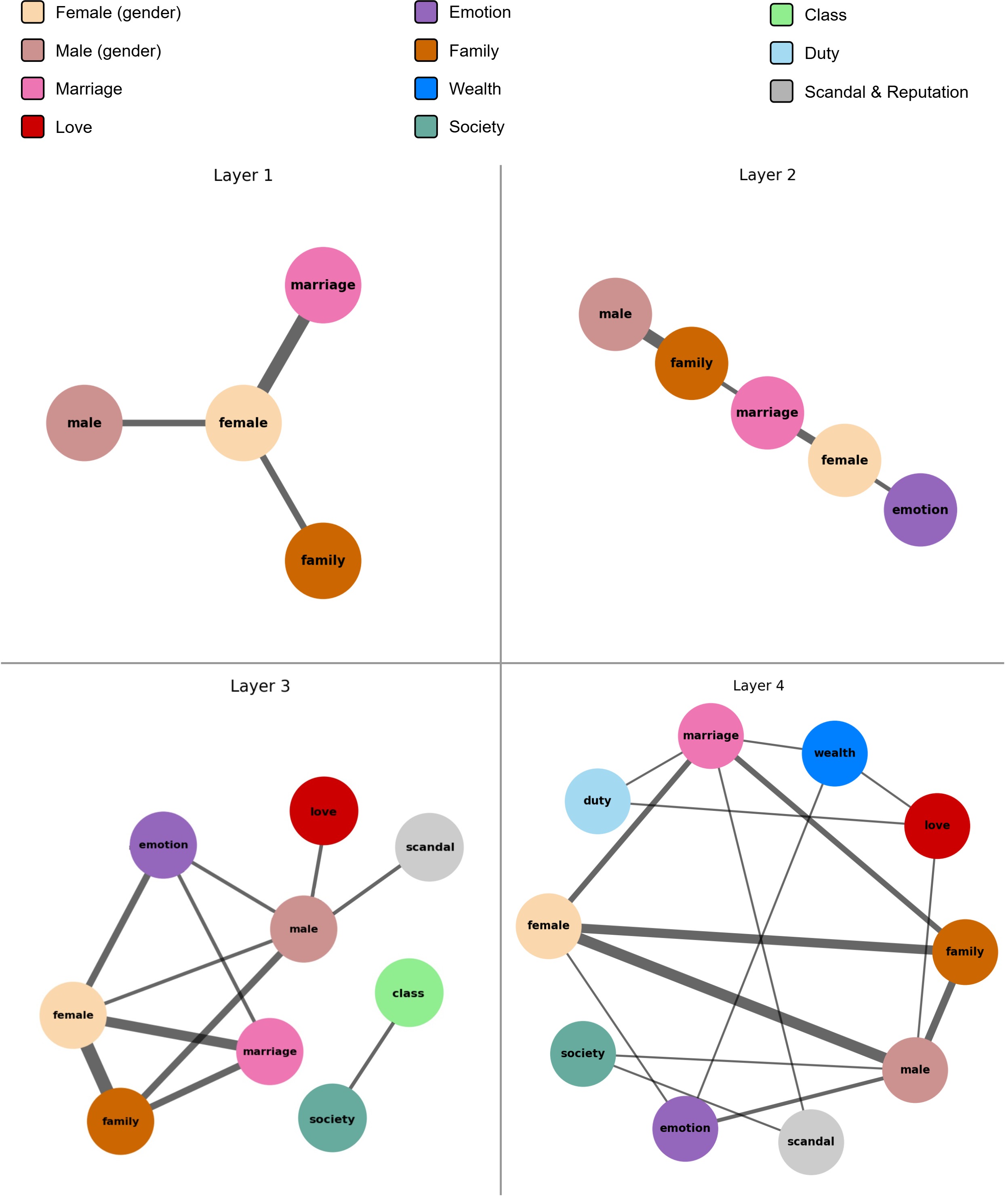}
    \vspace{3ex}
    \caption{Dual-themed neuron associations for Layers 1-4.}
    \label{fig:appendix_layers1to4}
\end{figure}

\begin{figure}[ht]
    \centering
    \includegraphics[width=0.8\textwidth]{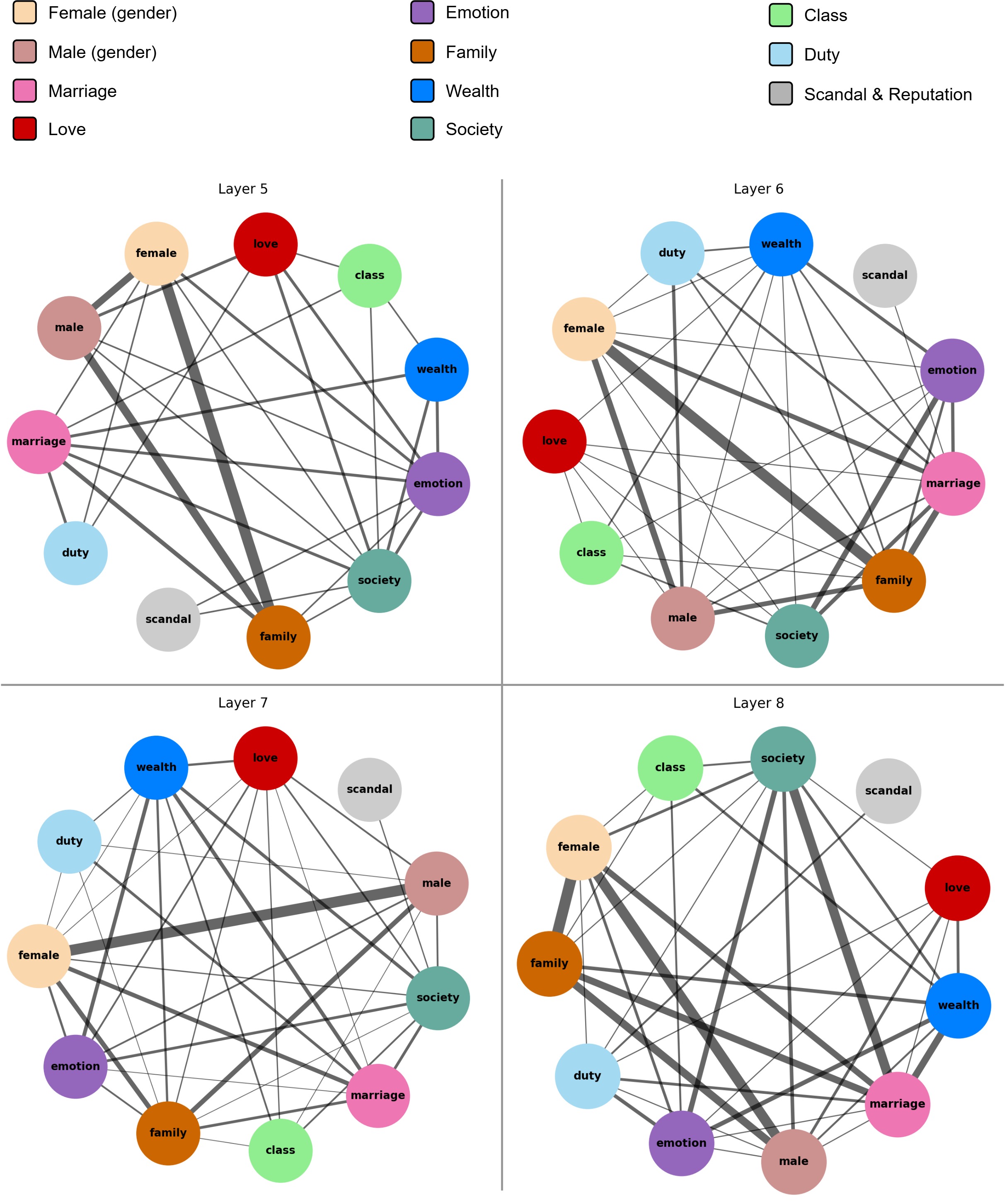}
    \vspace{3ex}
    \caption{Dual-themed neuron associations for Layers 5-8.}
    \label{fig:appendix_layers5to8}
\end{figure}

Figures~\ref{fig:appendix_layers1to4} and~\ref{fig:appendix_layers5to8} show the complete set of dual-theme neuron graphs for all eight layers. These visualizations complement the discussion in the main text, where Layers 1, 4, and 8 were highlighted as representative examples. Here, the full results are provided for reference.
\newline
\newline

\end{document}